\documentclass{article}

\usepackage{PRIMEarxiv}

\usepackage[utf8]{inputenc} 
\usepackage[T1]{fontenc}    
\usepackage{hyperref}       
\usepackage{url}            
\usepackage{booktabs}       
\usepackage{amsfonts}       
\usepackage{nicefrac}       
\usepackage{microtype}      
\usepackage{lipsum}
\usepackage{fancyhdr}       
\usepackage{graphicx}       
\graphicspath{{media/}}     
\usepackage{amsmath}

\title{When or What? Understanding Consumer Engagement on Digital Platforms
}

\author{
  Jingyi Wu \\
  Department of Linguistics \\
  Zhejiang University \\
  \texttt{jingyiwu@zju.edu.cn} \\
   \And
  Junying Liang \\
  Department of Linguistics \\
  Zhejiang University \\
  \texttt{jyleung@zju.edu.cn} \\
}

\begin{document}
\maketitle

\begin{abstract}
Understanding what drives popularity is critical in today’s digital service economy, where content creators compete for consumer attention. Prior studies have primarily emphasized the role of content features, yet creators often misjudge what audiences actually value. This study applies Latent Dirichlet Allocation (LDA) modeling to a large corpus of TED Talks, treating the platform as a case of digital service provision in which creators (speakers) and consumers (audiences) interact. By comparing the thematic supply of creators with the demand expressed in audience engagement, we identify persistent mismatches between producer offerings and consumer preferences. Our longitudinal analysis further reveals that temporal dynamics exert a stronger influence on consumer engagement than thematic content, suggesting that when content is delivered may matter more than what is delivered. These findings challenge the dominant assumption that content features are the primary drivers of popularity and highlight the importance of timing and contextual factors in shaping consumer responses. The results provide new insights into consumer attention dynamics on digital platforms and carry practical implications for marketers, platform managers, and content creators seeking to optimize audience engagement strategies.
\end{abstract}


\section{Introduction}
The proliferation of social media platforms has significantly increased individuals’ exposure to a vast array of information. Official statistics indicate that YouTube alone hosts over 51 million channels, yet only a small fraction reaches the milestone of amassing a million subscribers \cite{demandsage2024youtubestats}. Videos garnering over one million views typically generate more than \$2,000 in advertising revenue, with the highest-earning YouTuber securing an annual income of \$54 million. This scenario underscores the critical link between popularity and financial gain, highlighting the imperative for content creators to adeptly navigate and leverage prevailing popularity trends.\par

The question of how to achieve popularity has been a focal point of interest across various domains of communication, ranging from traditional information dissemination methods to modern mass media \cite{heath2007madetostick, archer2016bestseller, toubia2021shape, cheng2024rahg}. Notably, extensive research efforts have been dedicated to identifying factors that influence popularity trends on social media platforms. It has been established that real-world social networks exert a considerable influence on the dynamics of popularity \cite{zerubavel2015neural}, a finding that holds true in online social networks as well \cite{kwak2010twitter, weng2014memes}. Online platforms catalyze the cascading of information by enabling rapid dissemination through user interactions such as sharing, liking, and commenting. To decode the cascading effect within social networks, extensive studies have been conducted on their structures and dynamics \cite{cao2017deephawkes, das2024cascading}, leading to the development of models aimed at predicting popularity trends \cite{yang2024weibo}. The incorporation of temporal information to account for changes in popularity over time has also received considerable attention, with many studies employing multiple time-scales to capture evolving trends \cite{shang2021popularity}. Nevertheless, despite the importance of these factors, the intrinsic quality and the content itself are identified as the paramount determinants of popularity.\par

An experimental study \cite{figueiredo2014youtube} found that when extraneous factors are controlled, users’ perceptions of content tend to match its relative popularity, highlighting the strong influence of content quality on how popular it becomes. To delve deeper into the relationship between content and popularity, numerous studies have examined the dynamics of users’ attention economy in the context of social media content consumption. Popularity is widely recognized to be shaped by selective exposure \cite{stroud2010polarization, an2013fragmented, dubosar2025seeing}, with recent research providing empirical evidence for its critical role. This phenomenon is characterized by users’ propensity to direct their limited attention towards content that resonates with their personal interests or beliefs, thereby hastening the emergence of polarized groups and echo chambers \cite{delvicario2016echo, cinelli2021echo}. In other words, users exhibit a preference for information that aligns with their worldviews while disregarding contradictory data, with such individual preferences proving to be more influential than algorithmic recommendations in guiding their final content selections \cite{bakshy2015exposure, hartmann2025review}. Understanding user preferences is a paramount challenge for effective communication, with previous investigations shedding light on popularity trends through content analysis and exploration of audience interests in specific topics. Analysis of Twitter hashtags has revealed topic-based variations in information diffusion \cite{romero2011differences}, and a study on TED Talks utilized tags to delineate popularity trends among highly viewed videos \cite{johnson2023tedtrends}. Moreover, a comparative analysis across five media platforms demonstrated that the spread of information is intricately linked to the dynamics of user engagement with specific topics \cite{cinelli2020infodemic}, highlighting the critical role of topic selection in determining content popularity.\par

Aside from investigating user preferences, there is a mismatch between the content provided by creators and the expectations of users \cite{schmidt2017anatomy}. However, the preference gap between content creators and their audience has rarely been the subject of detailed research. Our research seeks to delve into and quantify these preferential differences. Communication can take many forms, with public speaking serving as a key medium for information dissemination and idea propagation, enabling social groups and individuals to achieve specific aims \cite{habermas2015between, garland2022hate}. The emergence of social media has revolutionized traditional communication methods, drawing 4.8 billion users \cite{chaffey2023socialmedia} to its vast and varied content. To broaden their reach, an increasing number of public speeches are being uploaded to platforms such as TED, podcasts, and various short video applications. The high volume and rapid turnover of content lessen the visibility of many messages \cite{vanbavel2021polarization}, while recommendation algorithms further shape which content is prioritized and surfaced to users \cite{haroon2023auditing, ibrahim2025tiktok}. As a result, speakers are often pushed to tailor their stories more strategically to fit algorithmic preferences and engage audiences effectively within tight time frames. Compared to other communication forms, online speeches provide richer content, enabling more precise inference of topics than the brief content typically found in short social media posts. Therefore, we focus on online public speeches to analyze the gap in topic preferences and investigate trends in popularity.\par

In this study, we conduct a comparative analysis of the thematic preferences between speakers and audiences across the years, focusing on uncovering the preference gaps. Our analysis exclusively utilizes TED Talks, a platform well-known for its broad spectrum of topics that span technology, entertainment, design, and beyond. TED’s repository, rich with thousands of speeches delivered by a diverse array of speakers, offers a unique dataset for our thematic exploration. Eschewing the direct use of hashtags for topic identification, we opt for Latent Dirichlet Allocation (LDA) modeling to conduct a more refined content analysis, aiming to unearth the latent thematic frameworks within TED Talks. This method, a statistical approach for analyzing text, is instrumental in extracting pivotal ideas, making it apt for our investigation into TED content \cite{blei2003lda, moghaddam2012aspectlda, dimaggio2013affinities, maier2021lda}. Our analysis encompasses 4,475 talks from 2006 to 2022, providing a substantial base for identifying trends and preferences.\par

Considering our initial goal to explore how thematic content and temporal variations contribute to the audience preference, we collected data on the number of videos per topic over the years and the view counts for each video. In our analysis, the frequency of videos on a specific topic serves as a metric for assessing speaker preferences. For audience preferences, we utilize the average view counts per topic, applying logarithmic transformation to process the data. We then examine yearly trends in these preferences and the discrepancies between speakers and audiences over time. By introducing the ‘difference index’ to measure the gap in preferences, our research finds that that both speakers’ and audiences’ topic interests are influenced by thematic attributes and temporal dynamics, albeit differently. This approach not only highlights the dynamic nature of preference but also quantifies the evolving gap between speakers and audiences across various topics.

\section{Methods}
\label{sec:headings}
\subsection{Dataset Construction}
In this study, we constructed a TED Talks Corpus by collecting all the transcripts of TED talks, consisting of 4,475 talks with 8,065,104 words in total, covering a wide array of topics. The data collection method adhered to the terms and conditions of the TED website (see: https://www.ted.com/about/our-organization/our-policies-terms/ted-com-terms-of-use). In this vein of inclusiveness, topic modeling could yield a persuasive clustering of TED talk content. These TED talks were published between 2006 and 2022, capturing the diachronic changes of the topic in TED talks over nearly two decades. Furthermore, we aggregated the view counts of each video as a quantitative measure of its popularity level.\par

\subsection{LDA Model Application for Topic Discovery}
The conventional qualitative coding of topics typically involves thematic analysis, in which researchers manually read through the data and assign codes that capture both explicit and implicit meanings in the text \cite{guest2012applied}. These codes are developed to represent emerging themes, and are applied to chunks of text through an interpretive process that requires substantial involvement from trained coders. This procedure can be highly time-consuming and depends heavily on coders’ subjective judgments, which may lead to variability in coding outcomes and reduce intercoder reliability. To overcome these limitations, this study adopts a computational method, Latent Dirichlet Allocation (LDA), which enables automated, scalable, and systematic identification of latent thematic structures, reducing human bias and enhancing replicability. This is a generative probabilistic model for collections of discrete data, most prominently used for topic modeling of text documents \cite{blei2003lda}. Compared with the conventional methods, the LDA model can discover latent topics that may elude manual inspection, effectively distilling high-dimensional text datasets into interpretable topic distributions. This scalability makes it adept at handling extensive corpora, ensuring that the topics identified are intrinsically linked to the corpus content and minimizing external biases \cite{griffiths2007topics, blei2009topic}. Furthermore, the LDA model is sufficiently flexible to extend and integrate with other models, showcasing its adaptability across various research domains, such as social media, digital libraries, and biology.\par

The foundational principle behind LDA is that documents can be viewed as mixtures of topics, with each topic being a distribution over a set of words \cite{blei2003lda}. In the generative process, for each document in the corpus, a distribution over topics is determined from a Dirichlet distribution. The assignment of topics to individual words is determined by evaluating the document’s proportional representation across a range of topics, and the topic with the highest probability is selected. Subsequently, specific words are chosen from the vocabulary that are predominantly associated with the identified topic \cite{blei2010probabilistic}. In this research, the LDA analysis was implemented using the scikit-learn package \cite{kramer2016scikitlearn}, a prominent machine learning library in Python. It is crucial to clarify that the LDA model generates distinct word clusters, commonly referred to as topics. However, the semantic label for each topic is not intrinsically determined by the algorithm. Instead, researchers interpret and name these topics based on the observation of high-frequency words within each word cluster. Moreover, a single document can be associated with multiple topics and LDA output provides the probability distribution of these topics for each document (e.g., 80\% \textit{Technology}, 20\% \textit{Internet}). In our analysis, we designate the topic with the highest probability as the predominant topic of the document.\par

Before processing the text documents in the LDA model, we preprocessed the data by retaining words with meaningful semantic value, specifically verbs, nouns, adjectives, and adverbs. To do so, we eliminated “stop” words—pronouns, prepositions and conjunctions—based on an exhaustive list from the \textit{nltk} Python package. Additionally, we retained lemmatized tokens longer than three characters were considered. This rigorous pre-processing ensures that the LDA model operates on a clean and semantically rich subset of the original texts. In the initial run of the model, the number of topics was set to 15 based on the perplexity, a measure used to evaluate the quality of probabilistic models, where a lower value indicates a better fit of the model to the data \cite{newman2009distributed}. However, the results were not satisfactory. Of the 15 derived topics, two exhibited poor word coherence, indicating ineffective topic clustering. Upon analysis of this iteration’s output, we observed that certain non-informative terms appeared with an anomalously high frequency, skewing the results. As a remedial measure, we further updated the stop word list by removing [‘kind’, ‘little’, ‘sort’, ‘show’, ‘maybe’, ‘like’, ‘great’, ‘whole’, ‘probably’, ‘part’, ‘point’, ‘number’, ‘second’, ‘line’, ‘still’]. In addition, without compromising the perplexity score, we made minor adjustments to the number of topics in an attempt to optimize the results.\par

\subsection{Data Normalization}
Normalization of the dataset is crucial in pre-processing and influences the results of statistical analysis \cite{mohamad2013standardization}. In our research, the view counts ranged from 253 to 75,072,123, which displayed pronounced heterogeneity, manifesting disparities across several orders of magnitude. Given the non-normal distribution of view counts with substantially larger values than the majority of data points (\textit{W}=.4379, \textit{p}<.001), the application of z-score standardization is inappropriate \cite{ali2014normalization}. Instead, logarithmic transformation is usually used to address skewed data and reduce the relative impact of extreme values or outliers. It can be useful if the data spans several orders of magnitude and reduces the relative impact of extreme values or outliers. In our research, we employed the natural logarithm transformation, using e (approximately 2.718) as the base. This approach not only addressed the aforementioned challenges but also simplified the interpretation, as the differences between log-transformed values can be directly related to multiplicative differences in the original scale \cite{lutkepohl2012log}. To compare the variations in view counts across topics over different years, we calculated the mean view counts.\par

\subsection{Statistical Analysis}
In our research, the video counts for a certain topic in a given year could be regarded as an indicator of speakers’ preference. If there is a bias in preference, the occurrence of various topics could be unevenly distributed over the years, which can be tested through the Chi-square ($\chi^2$) test of independence. By cross-tabulating the number of topics against discrete years, we calculated the $\chi^2$ statistic to evaluate the likelihood that any observed differences in video counts of each topic could be attributed to chance. A low \textit{p}-value (<.05) would indicate that the variations in video counts of each topic over time are statistically significant, suggesting a temporal shift in thematic focus across years. To further analyze the speakers’ preference in specific topics, it is essential to conduct post hoc tests \cite{franke2012chisquare}. Therefore, we calculated adjusted standardized residuals, which provide a measure of discrepancy between the observed and expected values, offering a perspective of interrelations between topics and years. To explore how thematic attributes and temporal factors contribute to speakers’ selection, a beta regression analysis was conducted to assess this influence. We performed the beta regression analysis using the \textit{betareg} package in R, with ‘topic’ and ‘year’ serving as independent variables and the proportion of video counts per topic as the dependent variable.\par

To quantify audience preference, we used the mean view counts of each topic for each year. Considering the non-normally distributed nature of mean view counts and the repeated measurements of topics across different years, we employed the Kruskal-Wallis H test to assess whether significant differences exist across topics and years. A beta regression was also conducted to examine how the variables ‘topic’ and ‘year’ contribute to view counts. Moreover, we presented the ranking of each topic every year to examine the temporal dynamics of topic popularity, gaining insights into trends and audience behavior over time.\par

To analyze the association between the preferences of speakers and audiences, we employed the Spearman’s correlation test. Building on the previous analysis of speaker and audience preferences, we focused on the preference gap between them. In our research, we introduced the ‘difference index’ to represent the preference gap. The difference index can be calculated as:

\begin{equation}
\text{Difference\ Index}_{\text{topic, year}} =
\left|
\frac{\text{Average\ View\ Count}_{\text{topic, year}}}
     {\text{Total\ View\ Counts}_{\text{year}}}
-
\frac{\text{Video\ Counts}_{\text{topic, year}}}
     {\text{Total\ Video\ Counts}_{\text{year}}}
\right|
\label{eq:difference_index_single}
\end{equation}

\begin{equation}
\text{Difference\ Index}_{\text{year}} = 
\sum_{\text{topics}} 
\left( 
\text{Difference\ Index}_{\text{topic, year}} 
\right)
\label{eq:difference_index_total}
\end{equation}

\begin{figure}[htbp]
    \centering
    \includegraphics[width=0.9\linewidth]{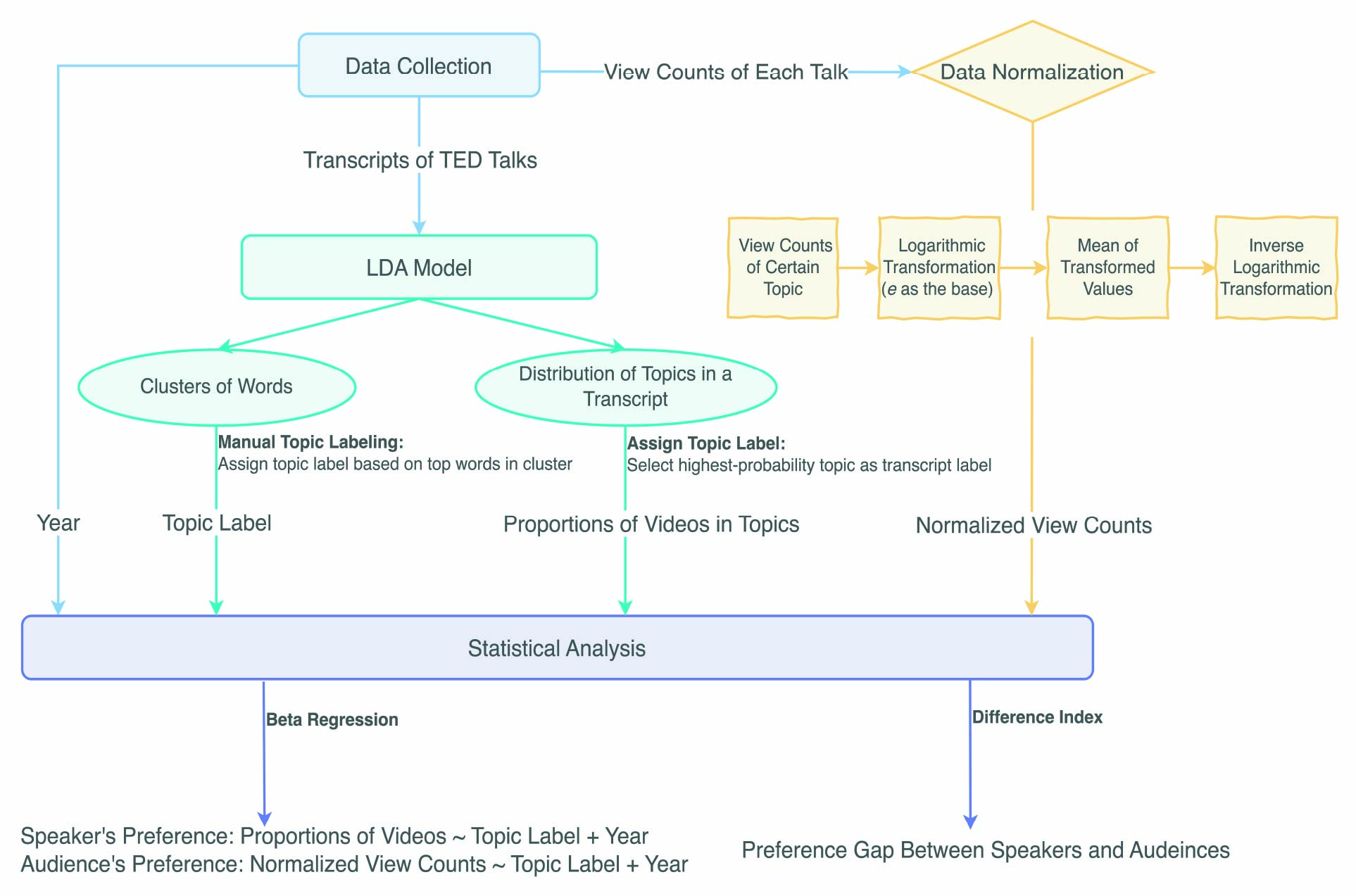}
    \caption{
        The flowchart of the study.
        This flowchart summarizes the data processing and analysis procedures, including topic modeling, data normalization, and statistical analysis.
    }
    \label{fig:flowchart}
\end{figure}

Using these equations, the annual difference index is computed by aggregating the absolute differences across topics, hence providing a measure to quantify the differences between speaker focus and audience engagement across topics for each year. Figure~\ref{fig:flowchart} presents the overall flowchart of the data processing and analysis procedure in this study.\par

\section{Results}
\subsection{Thematic preferences of speakers}
\begin{figure}[htbp]
    \centering
    \includegraphics[width=0.8\linewidth]{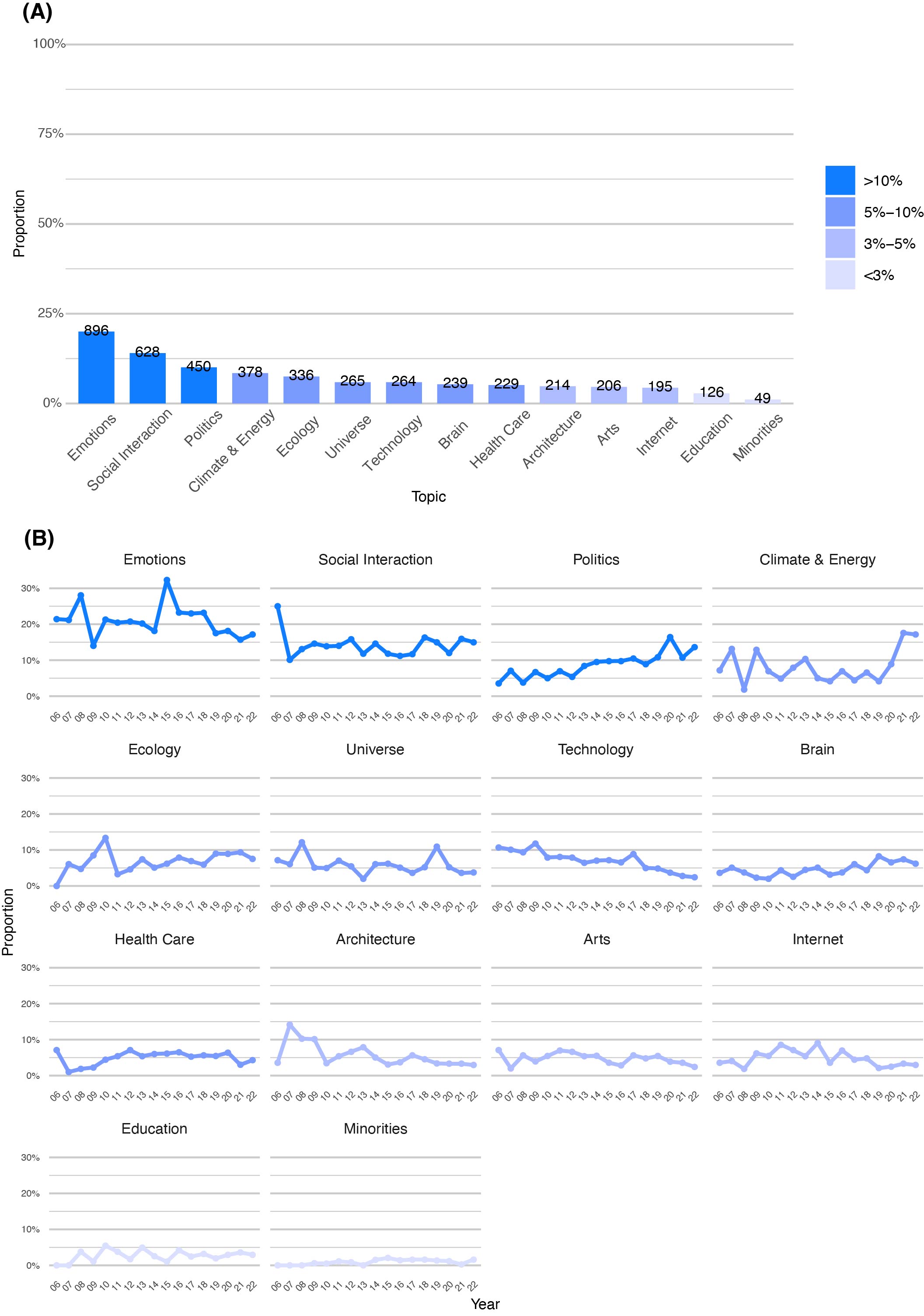}
    \caption{Distribution of 14 topics in TED talks. The figure provides a comprehensive overview of the distribution and temporal trends of 14 distinct topics. Each panel represents a unique aspect of the dataset analysed, illustrating the relative prominence and variation of topics across different years. Panel \textbf{(A)} illustrates the distribution of various topics within the dataset, using a bar chart where each bar represents the proportion of the total that each topic comprises. The bars are color-coded from dark blue to pale blue to indicate the size of the percentage each topic occupies. Panels \textbf{(B)} focus on single topic, detailing its annual percentage out of the total for each year covered in the study. Line graphs are utilized to depict the trend of each topic over time, highlighting fluctuations and patterns in their relative importance.}
    \label{fig:topic distribution}
\end{figure}

In this study, the video counts of each topic are regarded as an index to represent speakers’ preferences. The distribution of topics reflects the topic preferences of speakers. With the application of the LDA model, 4,475 TED talks were analyzed and yielded 14 distinct topics (see Figure~\ref{fig:topic distribution}A). Based on the frequent words, we summarized each topic and computed the respective proportions of videos corresponding to each topic (see Supplementary Table~\ref{tab:S1_topics_overview}). The following are identified topics: \textit{Emotions} (20.022\%), \textit{Social Interaction} (14.033\%), \textit{Politics} (10.056\%), \textit{Climate \& Energy} (8.447\%), \textit{Ecology} (7.508\%), \textit{Universe} (5.921\%), \textit{Technology} (5.899\%), \textit{Brain} (5.341\%), \textit{Health Care} (5.117\%), \textit{Architecture} (4.782\%), \textit{Arts} (4.603\%), \textit{Internet} (4.358\%), \textit{Education} (2.815\%), \textit{Minorities} (1.094\%).\par

Figure~\ref{fig:topic distribution} presents a comprehensive depiction of the distribution of topics in TED talks, along with the annual proportional variations of each topic as they evolved over the years. The distribution of videos across 14 identified topics is asymmetrical, with significant disparities in the corresponding proportions. Among these topics, the proportions of Emotions and Social Interaction are notably higher than others, comprising a quarter in total. By contrast, Minorities constitutes only a marginal fraction of the aggregate. This observation stands in stark contrast to conventional perceptions that minority-related issues are highly salient and widely discussed in public discourse, particularly given the apparent prevalence and heated debates surrounding minority topics in recent years \cite{williams2017ethics}. Overall, the spectrum of themes chosen by the speakers is extensive, exhibiting significant disparities in the degree of emphasis placed upon disparate topics.\par

For each topic, we observed significant variability in the yearly proportions, alongside clear divergence in the temporal patterns across various topics. We then assessed the potential association between topic selection and temporal distribution across years. The results of the Chi-square test ($\chi^2(208) = 537.922$, \textit{p} < .001) revealed a significant association between the variables, indicating that the preferences for topics are indeed influenced by the year. This suggests a temporally biased selection pattern, highlighting the dynamic interplay between time and thematic interests of speakers. To delve deeper into the interaction between years and topics, we employed residual analysis and discovered that some topics were sensitive to temporal factors, particularly \textit{Climate \& Energy} and Politics. In contrast, some topics remained consistently popular over time, such as Arts and Health Care (see Supplementary Figure~\ref{fig:S1_residuals}). Our findings align with a widely held notion that topics related to societal development and current affairs are more susceptible to temporal influences \cite{vliegenthart2011content, cinelli2020newsdiet, bento2020informationseeking}. Conversely, some topics that are universally relevant across societies—such as health and arts—demonstrated a marked resistance to temporal fluctuations, reflecting their enduring importance across different historical and cultural contexts \cite{boydstun2013making}. This phenomenon underscores the variability in how different topics respond to historical and cultural dynamics: some remain stable, while others are volatile.\par

To elucidate how thematic attributes and temporal factors contribute to speakers’ selection, a beta regression analysis was conducted with the ‘topic’ and ‘year’ serving as the independent variables and the proportion of videos assigned to each topic within each year as the dependent variable. In our analysis, we observed significant findings that enhance our understanding of speakers’ selection dynamics. The positive coefficients for certain topics, particularly Emotions ($\beta = 2.695$, SE = .214, \textit{p} < .001) and Social Interaction ($\beta = 2.231$, SE = .219, \textit{p} < .001), show that these topics account for a higher proportion of videos across years, suggesting their substantial influence on increasing the likelihood of speakers’ selection. However, Minorities ($\beta = -1.019$, SE = .303, \textit{p} < .001) exhibited negative association, indicating a deterrent effect on the selection process. These findings are consistent with the topic proportion shown in Figure~\ref{fig:topic distribution}A, underscoring the crucial role that thematic content plays in speakers’ selection. While the positive coefficients for the ‘year’ highlighted variability in speakers’ selection over time, they exhibited overall smaller values than those of the topic variables, suggesting that temporal factors also play a role, but less pronounced than thematic factors. According to the beta regression model (pseudo $R^2$= .361; for details, see Supplementary Table~\ref{tab:S2_beta_regression}), we can summarize that thematic content is slightly more influential than temporal influences in determining speakers’ preferences. This is likely because topic choice reflects speakers’ core interests and the primary messages they wish to convey, which are often personally or professionally motivated \cite{berger2012viral}. While temporal factors capture broader societal trends and evolving audience interests, they exert a more indirect influence by shaping the general salience of topics over time. Thus, although speakers may be responsive to changing social contexts, the intrinsic relevance of particular themes remains the dominant factor guiding their selection.\par

Despite the relatively minor role that temporal dynamics play in speakers’ selection, a meticulous analysis of the variations in topic prevalence across different years can significantly contribute to our comprehension of the selection process (see Figure~\ref{fig:topic distribution}B). None of these topics exhibited a clear and consistent upward or downward trajectory, indicating the temporal variability in predicting speakers’ preferences. Nonetheless, significant events within some specific years can still exert influence on speakers’ thematic selection, as indicated by distinctive peaks in these lines. For instance, the significant peak in Politics for 2020 corresponds to political unrest influenced by the global pandemic. Similarly, the peak in \textit{Climate \& Energy} in 2021 aligns with the year’s extreme heatwaves and natural disasters triggered by extreme weather. These peaks underscore the profound impact of global events on the selection of topics by speakers. In general, although the impact of temporal factors is not as pronounced as thematic attributes, specific events can indeed lead to noticeable shifts in speakers’ selection and preferences.\par

\subsection{Thematic preferences of audiences}
The view counts of videos represent the audience preferences and the popularity of topics. With data pre-processing (see Methods), we computed the average view counts of each topic from 2006 to 2022. To examine the differences in view counts among topics, a Kruskal-Wallis H test was conducted, which revealed substantial disparities across the 14 topics (\textit{H} = 30.118, \textit{df} = 13, \textit{p} = .005). This suggests that audience preferences for topics are not uniform, highlighting distinct variations in popularity levels.\par

\begin{figure}[htbp]
    \centering
    \includegraphics[width=0.8\linewidth]{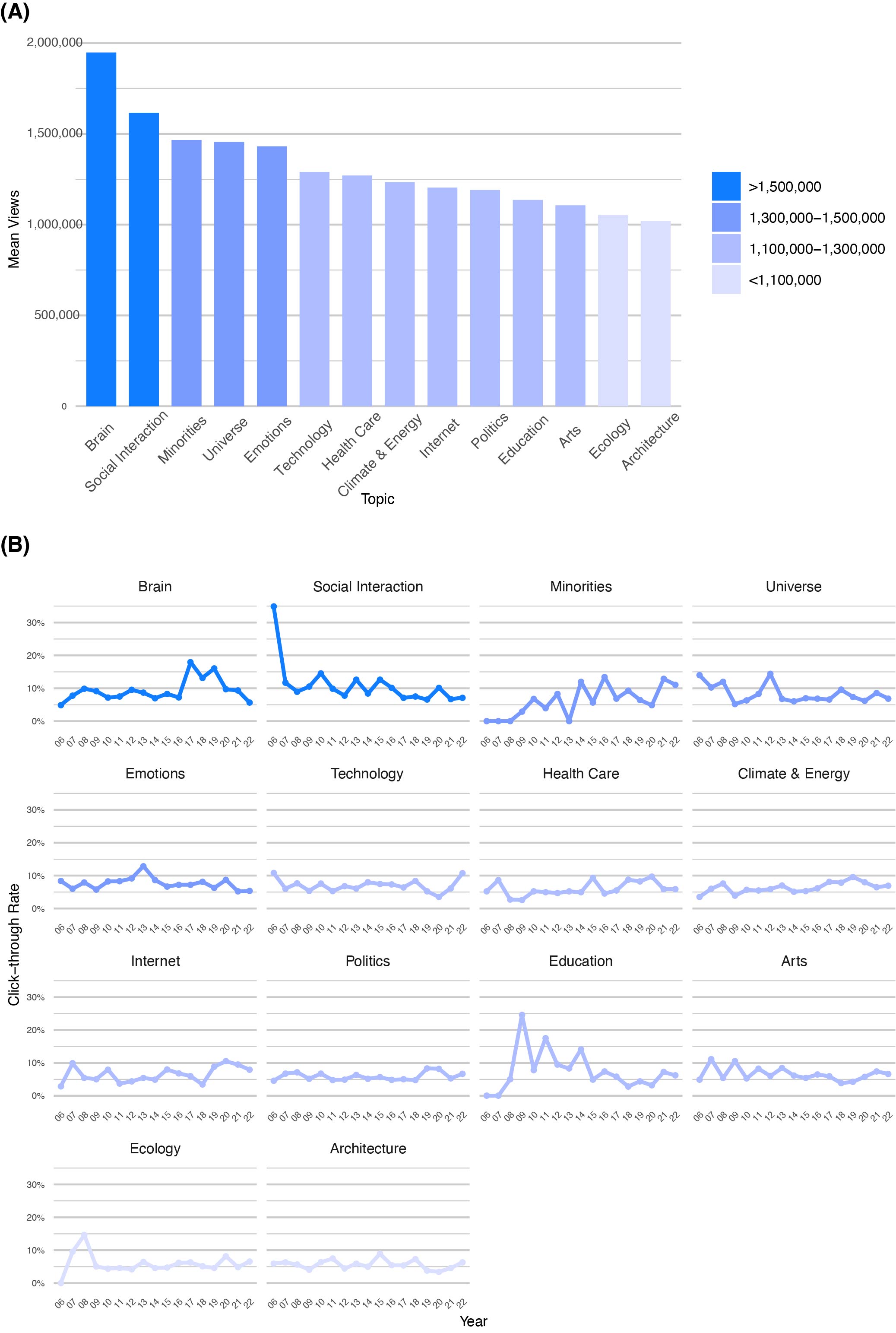}
    \caption{
        Popularity levels of each topic.
        The figure exhibits an overview of the mean view counts and temporal trends of 14 distinct topics, showing the shifting popularity levels of each topic over a span of 17 years. Panel \textbf{(A)} depicts the overall mean views of each topic, with each bar represents the logarithm transformed mean view counts that each topic possesses. Panel \textbf{(B)} show popularity trends of each topic across years, detailing the annual click-through rate for single topic. These line graphs display the dynamic changes in popularity of topics, highlighting fluctuation in audiences’ preferences.
    }
    \label{fig:popularity levels of topics}
\end{figure}

Following the initial identification of overall differences, post-hoc analyses using Dunn’s test were performed to pinpoint pairwise comparisons between topics. The results revealed that Social Interaction, Emotions, and Brain showed significant differences when compared across a multitude of topics, while no pronounced differences were observed among the remaining topics (see Supplementary Table~\ref{tab:S3_dunn}). To complement these findings, mean values of view counts were computed for each topic, facilitating a direct comparison of holistic popularity levels (see Figure~\ref{fig:popularity levels of topics}A). The top three topics with the highest mean view counts are Brain, Social Interaction and Minorities. Notably, the topic of Minorities occupied the smallest proportion among the 14 topics, but garnered high view counts disproportionate to its share. This divergence reflects a commonly observed gap between content supply and audience demand, where audience preferences favor underrepresented or emergent topics \cite{taneja2012measuring, webster2014marketplace}.\par

Similar to speakers’ preferences, audiences’ preferences are influenced by thematic content and temporal factors. The results of the beta regression model (pseudo $R^2=0.249$, for details see Supplementary Table S4) revealed that the majority of topics show positive association with view counts, particularly Social Interaction ($\beta =.975$, SE=.215, \textit{p} <.001) and Brain ($\beta = .933$, SE = .216, \textit{p} < .001), which have relatively higher coefficients than other topics. These results are identical to the ranking of mean views, indicating their popularity among audiences. However, the coefficient for Minorities is negative ($\beta = -.967$, SE=0.278, \textit{p} <.001), suggesting a seemingly adverse effect on view counts. This is contradictory to the popularity and positive impact on engagement of audiences observed from high mean views of Minorities. To clarify, the negative coefficient can be attributed to the absence of speeches on this topic in four years, resulting in zero clicks during these periods and skewing the coefficient for Minorities in this model. Moreover, the coefficients for temporal dynamics are all positive and exhibit overall higher values than the variable ‘topic,’ which is contrary to the results in the regression model of speakers’ selection. Higher coefficients for ‘year’ indicate that, for audiences, temporal factors exert a more potent influence on preferences than thematic attributes. This observation challenges our longstanding belief that content is most predictive of popularity, and highlights the audience’s heightened sensitivity to emerging trends and contextual salience \cite{garcia2017popularity, alshaabi2021storywrangler}. In fact, the impact of content on popularity may not be as pronounced as assumed.\par

To observe the variations in popularity levels of 14 topics across different years, we calculated the annual click-through rate for each topic over a span of 17 years (see Figure~\ref{fig:popularity levels of topics}B). These line graphs illustrate how thematic content and temporal shifts interact to influence views and contribute to understanding of audience engagement. The mean click-through rates for each topic exhibited considerable instability without a clear and consistent pattern across years. However, fluctuations in some topics during specific years are conducive to comprehending shifting audience interests. For example, the topic of Brain exhibited an overall higher click rate during the period from 2017 to 2019, indicative of an escalating interest in cognitive science. Between 2020 and 2023, Technology experienced a gradual increase in audience interest. The growing prominence of technology-related topics may reflect multiple driving forces. While ongoing technological advancements naturally stimulate sustained interest, the abrupt societal changes associated with the COVID-19 pandemic, such as widespread remote work and digital communication, may accelerate the visibility of technology in public discourse. Furthermore, the topic of Minorities has garnered increasing popularity with its relatively small share of videos, suggesting a disparity in interests between speakers and audiences. This disparity may reflect heightened public sensitivity to issues of diversity and social justice in recent years \cite{ince2017blm}. More broadly, our findings suggest a distinction between factors influencing speakers’ topic selection and audience preferences. The relatively stronger effect of temporal dynamics on audience attention suggests that audiences are more responsive to trending topics and societal shifts, which is contrary to the findings regarding speakers’ selection.\par

\begin{figure}[htbp]
    \centering
    \includegraphics[width=0.8\linewidth]{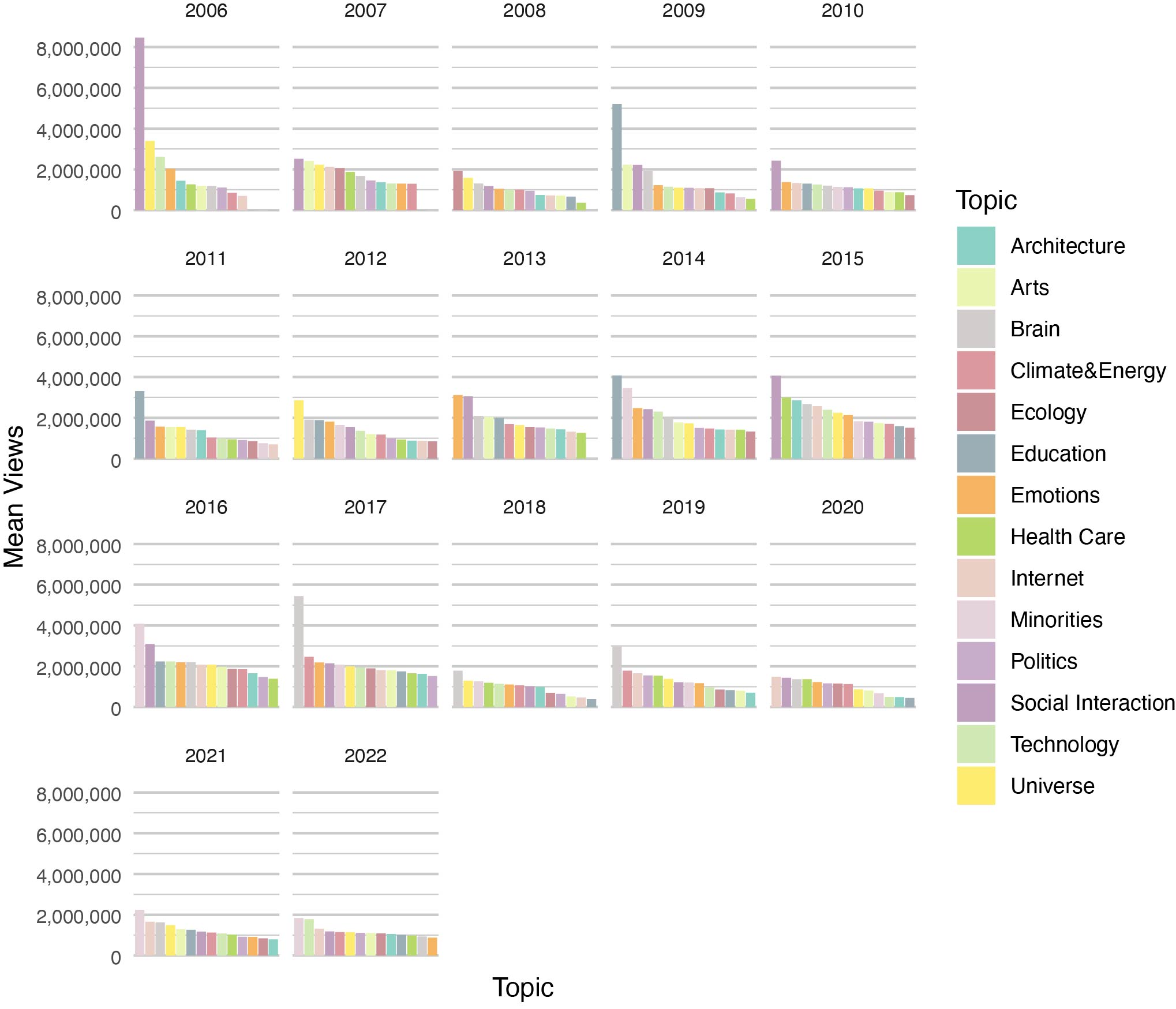}
    \caption{
        Rankings of average view counts of each topic across years.
        This figure displays the annual rankings of average view counts for each topic, where each coloured bar represents a different topic. The panels illustrate the rankings of all topics from 2006 to 2022.
    }
    \label{fig:Rankings of average view}
\end{figure}

To compare the shifting popularity levels of different thematic content, we ranked the average view counts of 14 topics in each year (see Figure~\ref{fig:Rankings of average view}). Despite the annual variations in rankings of mean view counts, certain topics consistently maintain high levels of popularity across multiple years. Among them, the topic of \textit{Social Interaction} topped the rankings for four years, with its highest average view counts reaching up to 8 million. \textit{Brain} consistently held the top position from 2017 to 2019, and maintained high view counts in subsequent years, a trend not observed in previous years. This trend indicates a growing audience interest in how human brains function, underscoring enduring curiosity about unsolved scientific mysteries \cite{rutjens2018attitudes}. The topic of \textit{Universe}, which can also be considered part of scientific mysteries, has received considerable attention, ranking first in 2012 and second in 2007, 2008, and 2018. Notably, Stephen Hawking delivered a TED Talk titled “Questioning the Universe,” which has accumulated over ten million views. Hawking’s prominent status as a globally recognized physicist and public intellectual likely contributed to the sustained audience interest in this topic \cite{chung2017parasocial}. Furthermore, his passing in 2018 reignited public attention to cosmological mysteries, illustrating how celebrity influence can amplify audience engagement and elevate the popularity of specific content. These incidents contribute to the high ranking of average view counts in \textit{Universe}, pinpointing the celebrity influence on audience preferences and popularity trends. In general, audience preferences for specific topics fluctuate over time, with no sustained long-term favoritism towards any topic. However, there are a handful of topics with high rankings across multiple years, highlighting the intrinsic appeal of their content. Furthermore, the popularity of certain topics is susceptible to specific events occurring in particular years, reflecting the dynamic nature of audience interests.\par

\subsection{Preference gap in topics between speakers and audiences}
We next analyzed the preference gap between speakers and audiences. A Spearman’s correlation test suggests a weak correlation between the preferences regarding various topics (\textit{r}(236) = .143, \textit{p} = .028), implying a divergence in their interests. To quantify this gap, we calculated the discrepancy between the annual proportion of videos for each topic and its average click-through rate within the same year. The absolute value of this discrepancy is defined as the ‘difference index’ in our study, representing the gap between speakers’ and audiences’ levels of interest in a specific topic for that year. The aggregate of values of difference indexes for all topics within a year represents the overall divergence in preferences between them.\par

Among the 14 topics, \textit{Arts}, \textit{Architecture} and \textit{Politics} exhibited relatively low and stable difference indices across years. \textit{Arts} and \textit{Architecture} are universal but significant topics, with their video counts maintaining a stable proportion across years (see Figure~\ref{fig:topic distribution}B). For such classic themes, audience interest tends to remain consistent over time, as these topics appeal to intrinsic aesthetic and cultural values that are less dependent on external circumstances \cite{walmsley2016audience}. Consequently, a significant deviation in preferences between speakers and audiences is generally not observed for these enduring topics. Compared with classic topics, Politics is sensitive to temporal factors and susceptible to current events. Because of this characteristic, when significant events occur, speakers and audiences are more likely to converge in their selections of this topic \cite{vaccari2015dualscreening}.\par

\begin{figure}[htbp]
    \centering
    \includegraphics[width=0.9\linewidth]{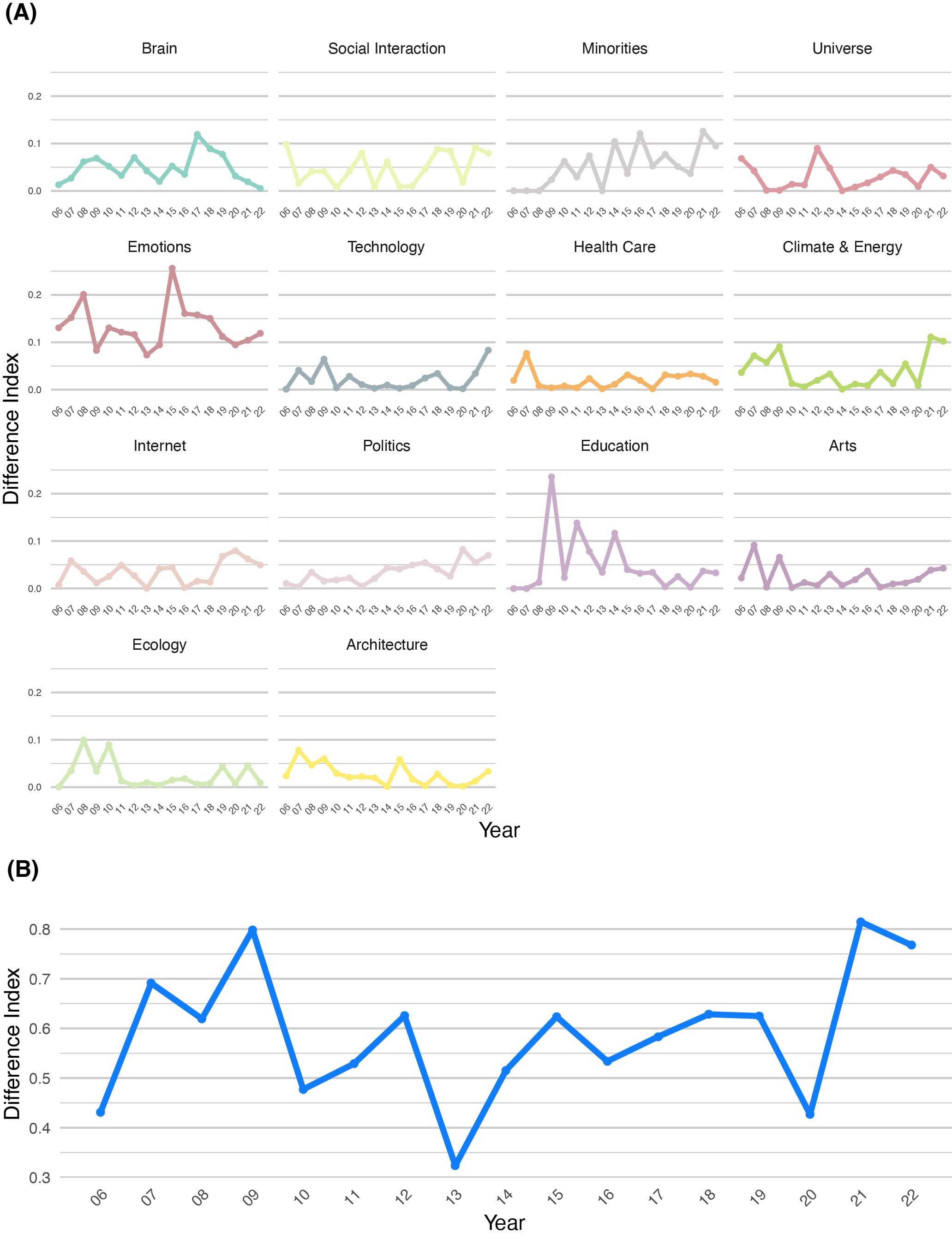}
    \caption{
        Overview of the difference index.
        This figure presents the annual difference indexes for each topic, along with the overall difference index per year. Panel \textbf{(A)} illustrate the evolving discrepancies in topic preferences between speakers and audiences over time, with each panel depicting the dynamic variance of a specific topic through the years. Panel \textbf{(B)} displays the annual difference index, which is the cumulative aggregation of difference indexes for all topics within a given year.
    }
    \label{fig:difference index}
\end{figure}

By contrast, the topic of Emotions exhibited overall high difference indices over time, indicating speakers’ misjudgment of its popularity among audiences (see Figure~\ref{fig:difference index}A). In certain years, some topics also showed large discrepancies in the preferences between speakers and audiences. For example, the differences in Brain were relatively large in the period between 2016 and 2019. This discrepancy can be attributed to the small proportion of videos, revealing incompatible interests between speakers and audiences in this topic. The topic of Minorities showed high difference indices across multiple years, with values exceeding 0.1 in three distinct years. In recent years, there has been increasing interest in this topic, but the published speeches have been scarce. This scarcity may result from the sensitivity of minority-related issues and a relatively small pool of speakers with sufficient expertise or willingness to address such topics \cite{lane2022inequality}. Therefore, the rarity of discourse contributes to the preference gap between speakers and audiences. Eventually, we computed the yearly difference index by adding up the annual differences (see Figure~\ref{fig:difference index}B). The resulting curve displayed no consistent upward or downward trajectory but instead fluctuated significantly across years, suggesting that preference gaps are dynamically shaped by evolving audience demands and the availability of speakers on specific topics \cite{scheufele2014political, hosseinmardi2021radical}.

\section{Discussion}
This study delves into how thematic content and temporal factors mutually contribute to the popularity trends of TED Talks and conducts a comparative analysis of topic preferences between speakers and audiences. According to results of the LDA model, 14 potential topics are unevenly distributed in TED talks, suggesting speakers’ preferences for certain topics. It is evident that more than one third of the topics fall under \textit{Emotions} and \textit{Social Interaction}, which are in a dimension of social psychology. This preference can be explained from two perspectives. From a utilitarian perspective, these topics can facilitate interpersonal proximity between communicators and induce empathetic resonance. Speakers share their personal experiences in their talks, which helps to build trust. From a perspective of contemporary relevance, social interactions are closely connected with individual well-being. It is confirmed that a harmonious relationship within the family \cite{crespo2011family, bell2018retrospective} and a positive role involvement in the workplace \cite{ryff2013eudaimonia, dahl2020plasticity} jointly enhance psychological well-being. Generally, perceived emotional support can mitigate depressive symptoms and dementia \cite{livingston2017dementia}, while social disconnectedness increases the risk of mental disorders \cite{santini2020disconnectedness}.\par

The topics of \textit{Universe}, \textit{Technology} and \textit{Brain} have relatively lower representation, hovering around 5\%. This scarcity can be attributed to the inherent accessibility challenges and complexity of these topics. They are in the domain of science communication, necessitating a thorough consideration of information disparities and cultural differences \cite{fischhoff2013sciences, fischhoff2014science2, valdezward2024communicators}. These topics require both speakers and audiences to have a robust foundation of scientific knowledge. However, the unequal access to information remains unsolved \cite{canfield2020critical}, probably leading to ineffective communication between speakers and audiences. Science communication is intertwined with cultural exchange \cite{medin2014cultural}, and a successful cultural transfer in a science-related talk can facilitate audience comprehension of the content. Experimental studies indicate that science communicators have less diversity in their cultural background and are taught many skills in storytelling instead of how to engage with diverse audiences \cite{dudo2021training}. Therefore, topics such as \textit{Universe}, \textit{Technology} and \textit{Brain} pose significant challenges for speakers due to the complexities inherent in science communication. Despite the increased attention towards minority groups in recent years, \textit{Minorities} ranks as the least addressed topic. This seemingly paradoxical occurrence can be attributed to the uniqueness and sensitivity of this topic, and the lack of prevailing public consensus. The reluctance of speakers to engage with the topic of \textit{Minorities} mirrors the ‘spiral of silence’ phenomenon resulting from fear of social isolation \cite{noelleneumann2004spiral, sohn2022spiral}. Consequently, a minimal number of speakers choose to discuss minorities in public forums. To recap, speakers consistently favor traditional and uncontroversial topics across years, demonstrating significant prudence in their public discourse.\par

The popularity of each topic is shaped by its thematic content and temporal factors, reflecting the diverse audience preferences towards various topics. Compared with speakers, audiences often gravitate towards more sensitive and science-related topics, such as \textit{Minorities}, \textit{Brain}, and \textit{Universe}. The popularity of these topics could be ascribed to curiosity sparked by an information gap \cite{pluck2011curiosity}. Cognitive interests thrive on uncertainty and prior studies have shown that blurring sensitive content can trigger information-seeking behavior \cite{bridgland2019triggerwarnings, simister2023informationgap}. In particular, sensitive or novel topics that are less frequently addressed can stimulate epistemic curiosity, motivating audiences to engage more deeply with such content \cite{gottlieb2018curiosity}. Therefore, the scarcity of readily available information on these topics motivates audiences to engage with these videos. Furthermore, temporal dynamics exert a more significant influence on audience engagement than thematic content, but this outcome is contrary to speakers’ preferences, where thematic content holds more influence. The diminished impact of thematic content on audiences’ preferences may suggest that content is not as pivotal as previously assumed.\par

The present study quantifies the topic preferences of speakers and audiences by video counts and view counts, and employs the difference index to measure the preference gap. Two notable preference gaps are detected: (1) the difference index exhibits significant variability across topics; (2) the difference index fluctuates across years with no particular pattern. The difference index of each topic varies significantly. \textit{Emotions}, \textit{Minorities}, and \textit{Brain} exhibit large difference indexes across years, and speakers overestimate the popularity of \textit{Emotions} while underestimating that of \textit{Minorities} and \textit{Brain}. Generally, successful social communication places great emphasis on who we address and a shared social identity could facilitate this process \cite{greenaway2015sharedidentity}. This ineffective communication could be traced to the psychological distinction between speakers and audiences. As we previously mentioned, speakers are conservative in choosing topics, but audiences make their choices to gratify their curiosity. Obviously, speakers lack this communicative ability. However, speakers make a successful prediction in \textit{Politics} and \textit{Health Care} with low difference indices. With further analysis, we find that the preference curves of speakers and audiences are relatively steady across years. Such stability could be ascribed to the fact that these two topics are significant and universal subjects associated with daily life and social welfare, so we always pay attention to them. As prior research has shown, topics related to political governance and public health consistently receive sustained public attention due to their personal and societal importance \cite{rainie2012pew, cai2021trends}.\par

After aggregating the yearly topic differences, we found that preferential differences between speakers and audiences always exist. The difference index curve fluctuates across years. It indicates that speakers fail to make some improvements to cater to the audience preferences and reveals the difficulty in predicting the popularity of the topics. Beyond the speakers’ inadequacy, the complexity of popularity trends could be attributed to some external factors. On the one hand, audiences are highly affected by social influence \cite{salganik2006inequality, bond2012socialinfluence}. To be specific, audiences are susceptible to the previous participants’ choices and show a tendency to agree \cite{centola2010spread}. Hence, videos with high view counts will continuously receive more views. On the other hand, the recommender system evaluates the users’ preferences and predicts their future likes and interests \cite{lu2012recommenders, zhang2021recommenderai}. Each user receives filtered information and individualized recommendations with a persuasive effect \cite{pathak2010recommender}, causing polarization of preferences between speakers and audiences.\par

\section{Conclusion}
In conclusion, this study offers a detailed analysis of topic features in TED Talks and proposes a quantitative approach to identify preference gaps between speakers and audiences by comparing their topic selections over time. While thematic content plays an important role in guiding speakers’ topic choices, temporal dynamics exert a more prominent influence on popularity among audiences.\par

Our findings reveal that both thematic content and temporal factors influence speakers' topic selection and audience preferences, though their effects differ between the two groups. For speakers, thematic content serves as an important driver of topic selection, while temporal dynamics play a comparatively smaller role. In contrast, temporal dynamics exert a stronger effect on audience popularity, indicating that content alone may not fully determine a speech’s success. These differing influences contribute to the preference gap between speakers and audiences, as captured by the difference index. Notably, topics such as Brain and Minorities exhibit larger preference gaps, with these disparities fluctuating over time without a consistent upward or downward trend.\par

Several limitations should be acknowledged. First, although we employed beta regression to assess the independent effects of thematic content and temporal dynamics, our analysis could not accommodate interaction terms due to model convergence issues caused by the large number of factor levels. However, we recognize that these factors may interact, as temporal shifts could influence the popularity of certain themes. Second, our analysis focused primarily on speech content, while external factors such as shares, likes, and audience engagement metrics may also affect popularity and merit inclusion in future research. Third, since our dataset is drawn from a single online platform, the generalizability of the observed trends may be limited and may not extend to other public speaking contexts.\par

This study also yields several implications. Initially, speakers should adjust their topics appropriately in line with current popular trends and place emphasis on disseminating scientific knowledge to bridge the information gap for the audience. There is a need to reduce the appearance of topics such as Social Interaction and Emotions, as these areas have become oversaturated and no longer provide novelty for the audience. Additionally, the selection of a topic does not necessarily predetermine the success of a public speech. Our research suggests that the impact of thematic content on popularity is less significant than previously assumed. Greater attention should be directed toward improving content delivery, particularly in terms of narrative structure and organization \cite{toubia2021shape}, which delineates a promising avenue for our future investigations. Furthermore, through quantifying the preference gap between speakers and audiences, we have observed that these disparities have not diminished over time, suggesting that many of our observations regarding popularity are largely stochastic. Integrating the outcomes of this study, we ascertain that for a more nuanced analysis of popularity, it is imperative to incorporate not only the content dimension but also to incorporate various external factors.\par

\bibliographystyle{unsrt}  
\bibliography{references}

\clearpage
\appendix

\section*{Supplementary Material}
\addcontentsline{toc}{section}{Supplementary Material}

\setcounter{table}{0}
\renewcommand{\thetable}{S\arabic{table}}
\setcounter{figure}{0}
\renewcommand{\thefigure}{S\arabic{figure}}

\begin{table}[htbp]
\centering
\caption{Overview of 14 Topics with LDA Model}
\label{tab:S1_topics_overview}
\renewcommand{\arraystretch}{1.2}
\begin{tabular}{p{3.2cm}p{3cm}p{8.5cm}}
\hline
\textbf{Topic Label} & \textbf{Proportion of Videos (\%)} & \textbf{Top 20 Most Frequent Words} \\
\hline
\textit{Emotions} & 20.02 & feel, love, never, live, home, talk, leave, walk, friend, moment, help, family, ever, story, always, keep, remember, turn, hand, away \\
\textit{Social Interaction} & 14.03 & talk, question, feel, idea, different, problem, person, important, experience, study, fact, answer, understand, example, sense, help, relationship, reason, self, course \\
\textit{Politics} & 10.06 & country, community, government, political, state, power, today, society, live, history, public, social, human, leader, system, place, century, create, believe, group \\
\textit{Climate \& Energy} & 8.45 & climate, percent, energy, money, dollar, country, company, business, problem, cost, global, economy, power, market, carbon, fuel, industry, economic, today, emission \\
\textit{Ecology} & 7.51 & water, food, animal, plant, specie, ocean, tree, fish, human, live, grow, forest, land, planet, nature, place, area, feed, protect, large \\
\textit{Universe} & 5.92 & light, planet, universe, science, space, energy, star, scientist, object, system, small, different, theory, form, solar, matter, dark, question, turn, idea \\
\textit{Technology} & 5.90 & technology, human, computer, game, machine, robot, video, play, build, future, create, move, system, able, device, real, design, different, problem, fast \\
\textit{Brain} & 5.34 & brain, cell, body, gene, sleep, human, memory, different, blood, system, cause, process, turn, control, form, information, activity, understand, study, able \\
\textit{Health Care} & 5.12 & health, patient, disease, cancer, drug, care, doctor, medical, treatment, virus, hospital, test, percent, risk, heart, system, treat, death, cause, blood \\
\textit{Architecture} & 4.78 & city, design, building, build, space, create, project, place, image, material, idea, house, wall, live, different, together, process, street, piece, structure \\
\textit{Arts} & 4.60 & story, write, book, word, language, music, sound, hear, read, play, talk, idea, film, speak, movie, listen, voice, artist, audience, name \\
\textit{Internet} & 4.36 & datum, information, company, Internet, network, system, technology, phone, online, medium, digital, share, build, social, example, able, access, service, help, tool \\
\textit{Education} & 2.82 & child, school, student, learn, family, parent, education, teacher, teach, help, young, community, class, high, program, mother, skill, study, baby, support \\
\textit{Minorities} & 1.09 & woman, black, girl, white, mother, color, young, violence, race, talk, baby, story, name, society, percent, body, bear, issue, wear, stand \\
\hline
\end{tabular}
\end{table}

\begin{figure}[htbp]
    \centering
    \includegraphics[width=0.9\linewidth]{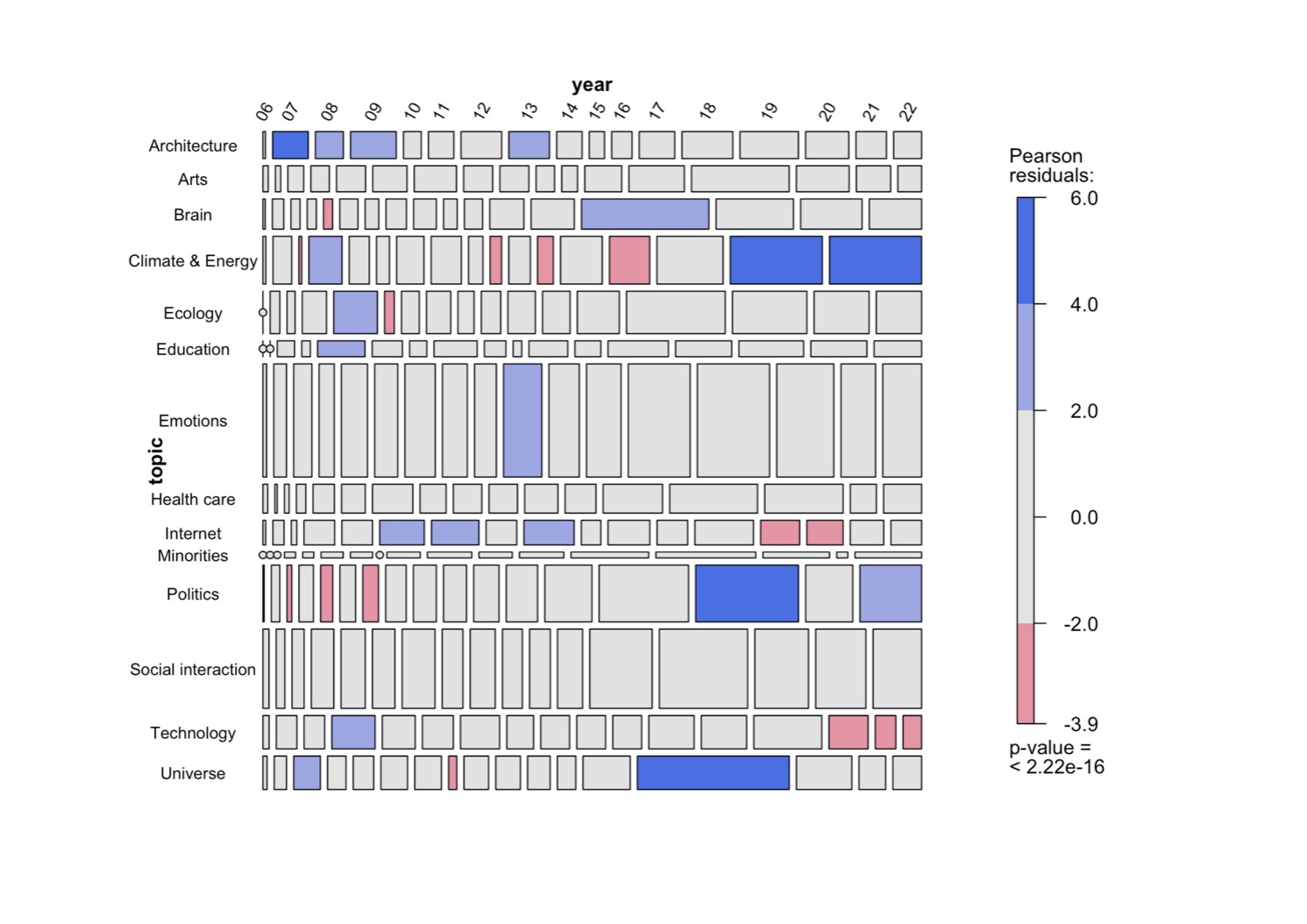}
    \label{fig:S1_residuals}
    \caption{
        Residual analysis of video counts.
        The mosaic figure describes how variables, \textit{topic} and \textit{year}, are correlated.
        Each tile represents the residual between the observed values and expected values.
        A residual between $-2$ and $2$ indicates that the two variables are independent under the null hypothesis of no association; otherwise, they are correlated.
        The analysis of residuals for video counts in topics such as \textit{Arts}, \textit{Health Care}, and \textit{Social Interaction} suggests that these topics’ popularity among speakers does not vary significantly with time.
        Conversely, \textit{Climate \& Energy} exhibited notable fluctuations, with higher video counts in 2009, 2021, and 2022, but lower counts in 2008, 2015, 2017, and 2019, reflecting the growing global concern over climate change and its impacts.
        Similarly, \textit{Politics} showed variable interest, particularly increasing in 2020 and 2022, possibly due to international political events.
        Unexpectedly, \textit{Technology} and \textit{Internet} displayed lower observed values in recent years—\textit{Internet} decreased in 2019–2020, and \textit{Technology} in 2020–2022.
        Despite rapid technological advancement and widespread adoption of large language models, this inconsistency may reveal a deviation of speakers’ focus from social trends.
        Moreover, coincident higher observed values in \textit{Brain} and \textit{Universe} in 2019 indicate mutual concerns regarding unsolved scientific mysteries.
    }
\end{figure}

\begin{table}[htbp]
\centering
\caption{Beta regression model of speakers’ preferences}
\label{tab:S2_beta_regression}
\begin{tabular}{lccccc}
\toprule
\textbf{Variables} & \textbf{Coefficient} & \textbf{Std. Error} & \textbf{\textit{z} score} & \textbf{\textit{p}-values} & \textbf{95\% Confidence Interval} \\
\midrule
(Intercept) & -4.8174 & 0.2650 & -18.179 & <0.001*** & (-5.3384, -4.2964) \\
\textit{Architecture} & 1.2252 & 0.2372 & 5.165 & <0.001*** & (0.7608, 1.6897) \\
\textit{Minorities} & -1.0190 & 0.3026 & -3.367 & <0.001*** & (-1.6116, -0.4264) \\
\textit{Health Care} & 1.1331 & 0.2396 & 4.729 & <0.001*** & (0.6631, 1.6032) \\
\textit{Universe} & 1.2835 & 0.2357 & 5.445 & <0.001*** & (0.8219, 1.7451) \\
\textit{Emotions} & 2.6952 & 0.2142 & 12.582 & <0.001*** & (2.2746, 3.1158) \\
\textit{Social Interaction} & 2.2305 & 0.2187 & 10.197 & <0.001*** & (1.8013, 2.6596) \\
\textit{Politics} & 1.7038 & 0.2267 & 7.515 & <0.001*** & (1.2596, 2.1481) \\
\textit{Technology} & 1.4519 & 0.2318 & 6.263 & <0.001*** & (0.9973, 1.9065) \\
\textit{Ecology} & 1.1328 & 0.2396 & 4.728 & <0.001*** & (0.6631, 1.6025) \\
\textit{Arts} & 1.1276 & 0.2397 & 4.703 & <0.001*** & (0.6599, 1.5953) \\
\textit{Climate \& Energy} & 1.5540 & 0.2296 & 6.767 & <0.001*** & (1.1043, 2.0037) \\
\textit{Internet} & 1.1240 & 0.2398 & 4.686 & <0.001*** & (0.6556, 1.5925) \\
\textit{Brain} & 1.0915 & 0.2407 & 4.535 & <0.001*** & (0.6201, 1.5630) \\
year2007 & 0.5311 & 0.2314 & 2.295 & 0.022* & (0.0780, 0.9841) \\
year2008 & 0.6966 & 0.2264 & 3.077 & 0.002** & (0.2536, 1.1396) \\
year2009 & 0.8423 & 0.2224 & 3.788 & <0.001*** & (0.4060, 1.2786) \\
year2010 & 0.8415 & 0.2224 & 3.784 & <0.001*** & (0.4052, 1.2779) \\
year2011 & 0.8914 & 0.2211 & 4.032 & <0.001*** & (0.4580, 1.3247) \\
year2012 & 0.8965 & 0.2210 & 4.057 & <0.001*** & (0.4631, 1.3299) \\
year2013 & 0.7972 & 0.2236 & 3.566 & <0.001*** & (0.3591, 1.2354) \\
year2014 & 0.9035 & 0.2208 & 4.092 & <0.001*** & (0.4706, 1.3364) \\
year2015 & 0.8620 & 0.2219 & 3.885 & <0.001*** & (0.4290, 1.2951) \\
year2016 & 0.8842 & 0.2213 & 3.996 & <0.001*** & (0.4512, 1.3171) \\
year2017 & 0.8944 & 0.2210 & 4.047 & <0.001*** & (0.4610, 1.3277) \\
year2018 & 0.9060 & 0.2207 & 4.104 & <0.001*** & (0.4728, 1.3393) \\
year2019 & 0.8527 & 0.2221 & 3.839 & <0.001*** & (0.4192, 1.2861) \\
year2020 & 0.8560 & 0.2220 & 3.855 & <0.001*** & (0.4225, 1.2895) \\
year2021 & 0.8141 & 0.2231 & 3.649 & <0.001*** & (0.3806, 1.2475) \\
year2022 & 0.8109 & 0.2232 & 3.633 & <0.001*** & (0.3774, 1.2444) \\
\bottomrule
\end{tabular}
\end{table}

\begin{table}[htbp]
\centering
\caption{Dunn’s tests results}
\label{tab:S3_dunn}

\begin{tabular}{lccccccc}
\toprule
\textbf{} & \textbf{\textit{Education}} & \textbf{\textit{Ecology}} & \textbf{\textit{Arts}} & \textbf{\textit{Climate \& Energy}} & \textbf{\textit{Internet}} & \textbf{\textit{Brain}} & \textbf{\textit{Architecture}} \\
\midrule
\textbf{\textit{Education}} & 1.000 &  &  &  &  &  &  \\
\textbf{\textit{Ecology}} & 1.000 & 1.000 &  &  &  &  &  \\
\textbf{\textit{Arts}} & 1.000 & 1.000 & 1.000 &  &  &  &  \\
\textbf{\textit{Climate \& Energy}} & 1.000 & 1.000 & 1.000 & 1.000 &  &  &  \\
\textbf{\textit{Internet}} & 1.000 & 1.000 & 1.000 & 1.000 & 1.000 &  &  \\
\textbf{\textit{Brain}} & 0.168 & <0.001*** & <0.001*** & <0.001*** & 0.002** & 1.000 &  \\
\textbf{\textit{Architecture}} & 1.000 & 1.000 & 1.000 & 1.000 & 1.000 & <0.001*** & 1.000 \\
\textbf{\textit{Minorities}} & 1.000 & 0.870 & 1.000 & 1.000 & 1.000 & 1.000 & 0.335 \\
\textbf{\textit{Health Care}} & 1.000 & 1.000 & 1.000 & 1.000 & 1.000 & 0.002** & 1.000 \\
\textbf{\textit{Universe}} & 1.000 & 0.082 & 1.000 & 1.000 & 1.000 & 0.578 & 0.018* \\
\textbf{\textit{Emotions}} & 1.000 & <0.001*** & 0.284 & 0.131 & 0.478 & 0.477 & <0.001*** \\
\textbf{\textit{Social Interaction}} & 0.080 & <0.001*** & <0.001*** & <0.001*** & <0.001*** & 1.000 & <0.001*** \\
\textbf{\textit{Politics}} & 1.000 & 1.000 & 1.000 & 1.000 & 1.000 & <0.001*** & 1.000 \\
\textbf{\textit{Technology}} & 1.000 & 1.000 & 1.000 & 1.000 & 1.000 & <0.001*** & 1.000 \\
\bottomrule
\end{tabular}

\vspace{1em}

\begin{tabular}{lccccccc}
\toprule
\textbf{} & \textbf{\textit{Minorities}} & \textbf{\textit{Health Care}} & \textbf{\textit{Universe}} & \textbf{\textit{Emotions}} & \textbf{\textit{Social Interaction}} & \textbf{\textit{Politics}} & \textbf{\textit{Technology}} \\
\midrule
\textbf{\textit{Education}} &  &  &  &  &  &  &  \\
\textbf{\textit{Ecology}} &  &  &  &  &  &  &  \\
\textbf{\textit{Arts}} &  &  &  &  &  &  &  \\
\textbf{\textit{Climate \& Energy}} &  &  &  &  &  &  &  \\
\textbf{\textit{Internet}} &  &  &  &  &  &  &  \\
\textbf{\textit{Brain}} &  &  &  &  &  &  &  \\
\textbf{\textit{Architecture}} &  &  &  &  &  &  &  \\
\textbf{\textit{Minorities}} & 1.000 &  &  &  &  &  &  \\
\textbf{\textit{Health Care}} & 1.000 & 1.000 &  &  &  &  &  \\
\textbf{\textit{Universe}} & 1.000 & 1.000 & 1.000 &  &  &  &  \\
\textbf{\textit{Emotions}} & 1.000 & 0.425 & 1.000 & 1.000 &  &  &  \\
\textbf{\textit{Social Interaction}} & 1.000 & <0.001*** & 0.226 & 0.044* & 1.000 &  &  \\
\textbf{\textit{Politics}} & 1.000 & 1.000 & 0.478 & 0.002 & <0.001*** & 1.000 &  \\
\textbf{\textit{Technology}} & 1.000 & 1.000 & 1.000 & 0.257 & <0.001*** & 1.000 & 1.000 \\
\bottomrule
\end{tabular}
\end{table}

\end{document}